%% file: sample-sigconf.tex
\documentclass[dvipsnames,format=sigconf,anonymous=false,review=false]{acmart}

\usepackage{colortbl}
\usepackage{color}
\usepackage{graphicx}
\usepackage{multirow}
\usepackage{float}
\definecolor{lightgray}{gray}{0.9}
\AtBeginDocument{%
  \providecommand\BibTeX{{%
    \normalfont B\kern-0.5em{\scshape i\kern-0.25em b}\kern-0.8em\TeX}}}

\copyrightyear{2023}
\acmYear{2023}
\setcopyright{rightsretained}
\acmConference[GECCO '23 Companion]{Genetic and Evolutionary Computation
Conference Companion}{July 15--19, 2023}{Lisbon, Portugal}
\acmBooktitle{Genetic and Evolutionary Computation Conference Companion
(GECCO '23 Companion), July 15--19, 2023, Lisbon,
Portugal}\acmDOI{10.1145/3583133.3590751}
\acmISBN{979-8-4007-0120-7/23/07}

\usepackage{algorithm}
\usepackage{algpseudocode}
\usepackage{diagbox}

\algrenewcommand\algorithmicrequire{\textbf{Data:}}
\algrenewcommand\algorithmicensure{\textbf{Result:}}

\usepackage{caption}
\usepackage{subcaption}
\usepackage{graphicx}

\begin{document}

\title[]{A Static Analysis of Informed Down-Samples}
%

\author{Ryan Boldi}
\affiliation{%
  \institution{University of Massachusetts Amherst}
  \country{}
}
\email{rbahlousbold@umass.edu}

\author{Alexander Lalejini}
\affiliation{%
  \institution{Grand Valley State University}
  \country{}
}
\email{lalejina@gvsu.edu}

\author{Thomas Helmuth}
\affiliation{%
  \institution{Hamilton College}
  \country{}
}
\email{thelmuth@hamilton.edu}

\author{Lee Spector}
\affiliation{
  \institution{Amherst College}
  \country{}
}
\email{lspector@amherst.edu}

\renewcommand{\shortauthors}{Anonymous}

\begin{abstract}
    We present an analysis of the loss of population-level test coverage induced by different down-sampling strategies when combined with lexicase selection. We study recorded populations from the first generation of genetic programming runs, as well as entirely synthetic populations. 
    Our findings verify the hypothesis that informed down-sampling better maintains population-level test coverage when compared to random down-sampling.
    Additionally, we show that both forms of down-sampling cause greater test coverage loss than standard lexicase selection with no down-sampling.
    However, given more information about the population, we found that informed down-sampling can further reduce its test coverage loss.
    We also recommend wider adoption of the static population analyses we present in this work.
\end{abstract}




\maketitle


\section{Introduction}

Lexicase selection is designed for use on test-based problems ~\citep{assessment_spector_2012, Helmuth_solving_2015}; that is, problems where a candidate solution's quality is assessed using a training set comprising input-output examples that specify correct behavior.
Indeed, lexicase selection has been shown to be effective across a wide range of problem domains in evolutionary computing~\citep{helmuth_applying_2022, stanton_lexicase_2022, Ding2022optimizing, lalejini_artificial_2022}. 
However, evaluating a population on the full set of training cases can be expensive when using a large population or when evaluations are computationally intensive.
Given fixed computational resources, this evaluation cost can severely constrain the number of generations that we can run an evolutionary search.
Down-sampling the training set reduces the per-generation cost of evaluation, allowing practitioners to run deeper evolutionary searches with equivalent computing resources.

By enabling deeper evolutionary searches for a fixed computational budget, both informed down-sampling and random down-sampling have been shown to improve problem-solving success when combined with lexicase selection \citep{hernandez_random_2019, Helmuth2021benefits, boldi_tournDS_2023}.
These improvements are often substantial, as down-sampled lexicase has been shown to be state-of-the-art on program synthesis benchmarks \citep{helmuth_applying_2022}. 
Random down-sampling uses a random subset of training cases for selection each generation \citep{moore2017lex, lasarczyk_dynamic_2004, hernandez_random_2019}.
Random down-sampling, however, can omit important training cases from the sample (e.g., a training case that represents edge-case behavior).
Such omissions result in generations where parent selection is not conditioned on individuals' ability to solve these training cases, which can result in the loss of important diversity and impede problem-solving success \citep{hernandez_exploration_2022}.


Informed down-sampling \citep{boldi_informed_2023} addresses this drawback of random down-sampling by using run-time population statistics to construct down-samples containing training cases that are more distinct from one another. 
This helps to prioritize the inclusion of training cases that only a subset of the population solve while reducing the number of synonymous cases\footnote{Synonymous or nearly synonymous cases are solved by similar groups of individuals in the population. Therefore, the population's performance on these test cases is strongly correlated.} in the down-sample.
Indeed, informed down-sampling has been shown to increase the success rate of GP runs even further than random down-sampling \citep{boldi_informed_2023}.
The authors of this study hypothesized that informed down-sampling's edge over random down-sampling stems from its ability to better maintain distinct candidate solutions that specialize on different subsets of training cases, potentially improving the maintenance of important genetic building blocks for solving the problem.

Previous work has shown that down-sampling can cause a reduction in a population's test coverage when using lexicase selection~\cite{hernandez_exploration_2022}.
A population's \textit{test coverage} measures the number of training cases solved by at least one individual in that population.
In this study, we investigate the claim that informed down-sampling mitigates test coverage loss relative to random down-sampling. 
To do so, we analyzed test coverage loss when using different down-sampling methods with lexicase selection on two types of populations: recorded populations from the first generation of real program synthesis runs and synthetic populations configured to exemplify potential differences in test coverage loss. 
For each type of population, we compared the test coverage loss after applying lexicase selection under different down-sampling conditions, including two forms of informed down-sampling, random down-sampling, and no down-sampling (standard lexicase). 
Overall, our results verify that random down-sampling is more susceptible to causing test coverage loss than informed down-sampling. 
Our results are also consistent with previous studies showing that down-sampling can sometimes dramatically increase test coverage loss, while standard lexicase with no down-sampling loses little to no test coverage. 

\section{Methods}

\subsection{Selection methods}

We analyze test coverage loss for four variants of lexicase selection, each configured with a different down-sampling technique: random down-sampled lexicase (Rand), informed down-sampled lexicase (IDS), full-information informed down-sampled lexicase (Full-IDS), and standard lexicase with no down-sampling (Lex). 

\paragraph{Lexicase selection.} 
We use standard lexicase selection as the base parent selection algorithm for our analyses. 
To select a parent, lexicase selection shuffles the training cases into a random order and initially considers the entire population as candidates for selection.
The pool of candidates is filtered down by each training case in sequence, maintaining only the individuals that are elite (among the remaining candidates) on the current training case before moving on to the next training case. 
If only one candidate remains after filtering on a training case, that individual is selected. 
If all training cases are used and multiple individuals remain as candidates, a candidate is selected at random. 
More detailed descriptions of the lexicase selection procedure can be found in \citep{Helmuth_solving_2015}.

\paragraph{Random down-sampled lexicase selection.}
Random down-sampled lexicase creates a random down-sample of training cases each generation and then applies standard lexicase selection using only the down-sampled set of training cases \citep{hernandez_random_2019}.

\paragraph{Informed down-sampled lexicase selection.}
Informed down-sampl\-ing evaluates a random subset of the population on the full training set and then estimates the distance between each pair of training cases based on the distribution of individuals that solve them.  
We sample 1\% of the population to evaluate on the full training set. 
For example, two training cases that are solved by the same set of individuals have a distance of zero and are functionally synonymous, whereas two training cases solved by complementary groups of individuals are distant from one another and considered mutually informative.
These pairwise distances between training cases are used to iteratively construct a down-sample by first including a random training case in the down-sample and then by repeatedly selecting the training case that has the maximum minimum distance from all training cases already included in the down-sample.
See \citep{boldi_informed_2023} for a detailed description of the informed down-sampling procedure.
In informed down-sampled lexicase, the informed down-sampling procedure is used to create a subset of the training set, and then standard lexicase selection is applied using this subset. 

\paragraph{Full-information informed down-sampled lexicase selection.}
As in previous work \citep{boldi_informed_2023}, we use full-information informed down-sampled lexicase selection as a control.
This method first evaluates the entire population on the full training set, and then applies the informed down-sampling procedure to down-sample the training set.
Next, standard lexicase selection chooses parents using the down-sampled training set.
By using the full population to calculate distances between all pairs of training cases, this method does not reduce the number of per-generation evaluations needed to select parents, and thus, we do not recommend using this technique in practice. 
This technique, however, is a useful control, as it provides a best-case scenario for informed down-sampling; that is, it allows us to measure informed down-sampling's potential assuming we had access to perfect information about the population's performance on all training cases.

\subsection{Test coverage analysis}


We define a population's test coverage as the number of training cases solved by at least one individual (i.e., the number of training cases ``covered'' by a population).
Test coverage can indicate how much of a problem a population collectively solves.
A \textit{decrease} in test coverage from one generation to the next indicates the loss of some functionality required to solve the overall problem, which can be counter-productive for evolutionary search.
A population's test coverage can increase only as a result of mutation or crossover. 
In the absence of variation operators, parent selection can only contribute to a loss of test coverage (e.g., if all individuals that solve a particular training case are not selected).

Here, we measure \textit{test coverage loss} as the difference in test coverage between the original population and the selected parents.
We compare the test coverage loss that results from applying each of the four selection methods described above to a population.
In all experiments, we selected a number of parents equal to the original population size of 1,000. 

\subsubsection{Recorded population analyses}

\input{figs/recorded-pop-fig.tex}

We recorded the populations from the first generation of 80 independent genetic programming runs: 40 attempting to solve the Count Odds program synthesis problem \citep{helmuth_general_2015}, and 40 attempting to solve the Fizz Buzz program synthesis problem \citep{helmuth_psb2_2021}.
We performed selection ten times on each population using full-information informed down-sampling, informed down-sampling, random down-sampling, and no down-sampling (standard lexicase).
We repeated the analysis at three different down-sampling levels: 0.05, 0.1 and 0.2, meaning 5\%, 10\% or 20\% of the entire training set (200 training cases) is included in each of the down-samples.
For each down-sampling level, we compared each selection condition's test coverage loss.
To differentiate between these distributions, we first performed a Kruskal-Wallis test to detect significant differences across conditions ($\alpha < 0.05$), and if so, we applied a Wilcoxon rank-sum test to distinguish between pairs of selection conditions using a Bonferroni correction for multiple comparisons. 

These recorded populations represent realistic scenarios encountered early on during an evolutionary search.
While realistic, these populations have a large amount of variance in the distribution of individuals able to solve different test cases, motivating our synthetic population analyses.

\subsubsection{Synthetic population analyses}

To complement the recorded populations, we manually created synthetic populations with certain population-level properties by creating a set of 1,000 error vectors representing individuals. We used these synthetic populations to exemplify differences in a selection scheme's capacity to preserve population-level test coverage.
We evenly distributed individuals among 10 mutually exclusive niches. 
Each niche was associated with 20 training cases (out of the 200 total), and all 100 individuals assigned to a niche passed all 20 of the associated training cases while failing all others.
Therefore, test coverage could decrease only when every individual in a niche failed to be selected.

For our synthetic population analyses, we limited comparisons to informed down-sampling, random down-sampling, and no down-sampling (standard lexicase).
As in our recorded population analyses, we repeated this analysis at three different down-sampling levels: 0.05, 0.1 and 0.2.
For each configuration, we ran 20 replicate analyses, each with a unique random number seed.
Unlike our recorded population analyses, we applied 300 rounds of selection in each replicate analysis. 
For each round, we ran the treatment-specific selection method to choose parents, and then we copied those parents to create a ``new'' population to be used in the next round. 
No evolution occurred in these analyses, as we did not mutate individuals at any point. 
By applying additional rounds of selection, we can further isolate how stably a selection procedure maintains high test coverage.

\section{Results}


Figure~\ref{fig:co} shows the test coverage loss for performing selection on a random population initialized to solve the Count Odds and Fizz Buzz problems. 
We see that, for small down-samples ($r{=}0.05$), IDS has a significantly lower test coverage loss than random down-sampling (Wilcoxon rank-sum, $p{<} 0.005$). Lexicase selection maintains all of the test coverage (loss of zero) for all runs and configurations. Full information IDS significantly outperforms both random and informed down-sampling on all problems and down-sample sizes (Wilcoxon rank-sum, $p{<}10^{-4}$). In fact, for the Fizz Buzz problem, Full-IDS maintains test coverage as well as lexicase selection does, despite only having 10, 20, or 40 training cases in the sample.


Figure~\ref{fig:synthetic-pops} shows the test coverage loss from our synthetic population analyses.
Overall, our data show that down-sampling substantially increases test coverage loss relative to standard lexicase with no down-sampling.
Strikingly, standard lexicase selection maintains full test coverage for all 300 selection rounds across all replicates.
Consistent with the recorded population analyses, random sampling causes more test coverage loss than informed down-sampling. 
As expected, as the down-sample size increases, the rate of test coverage loss slows for both informed and random down-sampling, as more niches are likely to be included in the down-sample.

\input{figs/synthetic-pop-fig.tex}

Overall, our analyses support the hypothesis that informed down-sampling better maintains population-level test coverage as compared to random down-sampling.
Interestingly, the Full-IDS results from the recorded population analysis suggest that increasing the parent sample size ($\rho$) can reduce test coverage loss at the cost of needing to evaluate more individuals on the entire training set.
We also recommend wider adoption of the kinds of static population analyses performed in this work.
These types of analyses can be applied to any selection procedure, and can help us to better understand how selection schemes work in the context of different population structures.



\begin{acks}
The authors would like to thank Anil Saini, Charles Ofria, Dominik Sobania, Edward Pantridge, Franz Rothlauf, Li Ding, Martin Briesch, Nic McPhee, and the members of the PUSH Lab at Amherst College for discussions that helped shape this work. This material is based upon work supported by the National Science Foundation under Grant No. 2117377. Any opinions, findings, and conclusions or recommendations expressed in this publication are those of the authors and do not necessarily reflect the views of the National Science Foundation. This work was performed in part using high performance computing equipment obtained under a grant from the Collaborative R\&D Fund managed by the Massachusetts Technology Collaborative.
\end{acks}

\bibliographystyle{ACM-Reference-Format}
\bibliography{sample-base}

\end{document}

%% file: figs/recorded-pop-fig.tex


\begin{figure*}[ht]
\caption{Test coverage loss for each of full information informed down-sampling (Full-IDS), sparse information informed down-sampling (IDS), random down-sampling (Rand) and standard lexicase selection at the $r=0.05$, $0.1$ and $0.2$ down-sampling rates. Panels a - c and d - f are results from random populations generated for solving the Count Odds and Fizz Buzz problems, respectively.
Kruskal-Wallis tests for all down-sample configurations were statistically significant ($p < 0.03$).
}
     \centering
     \begin{subfigure}[b]{0.9\textwidth}
     \begin{subfigure}[b]{0.3\textwidth}
         \centering
         \subcaption{Count Odds, $r=0.05$}
         \includegraphics[width=\textwidth]{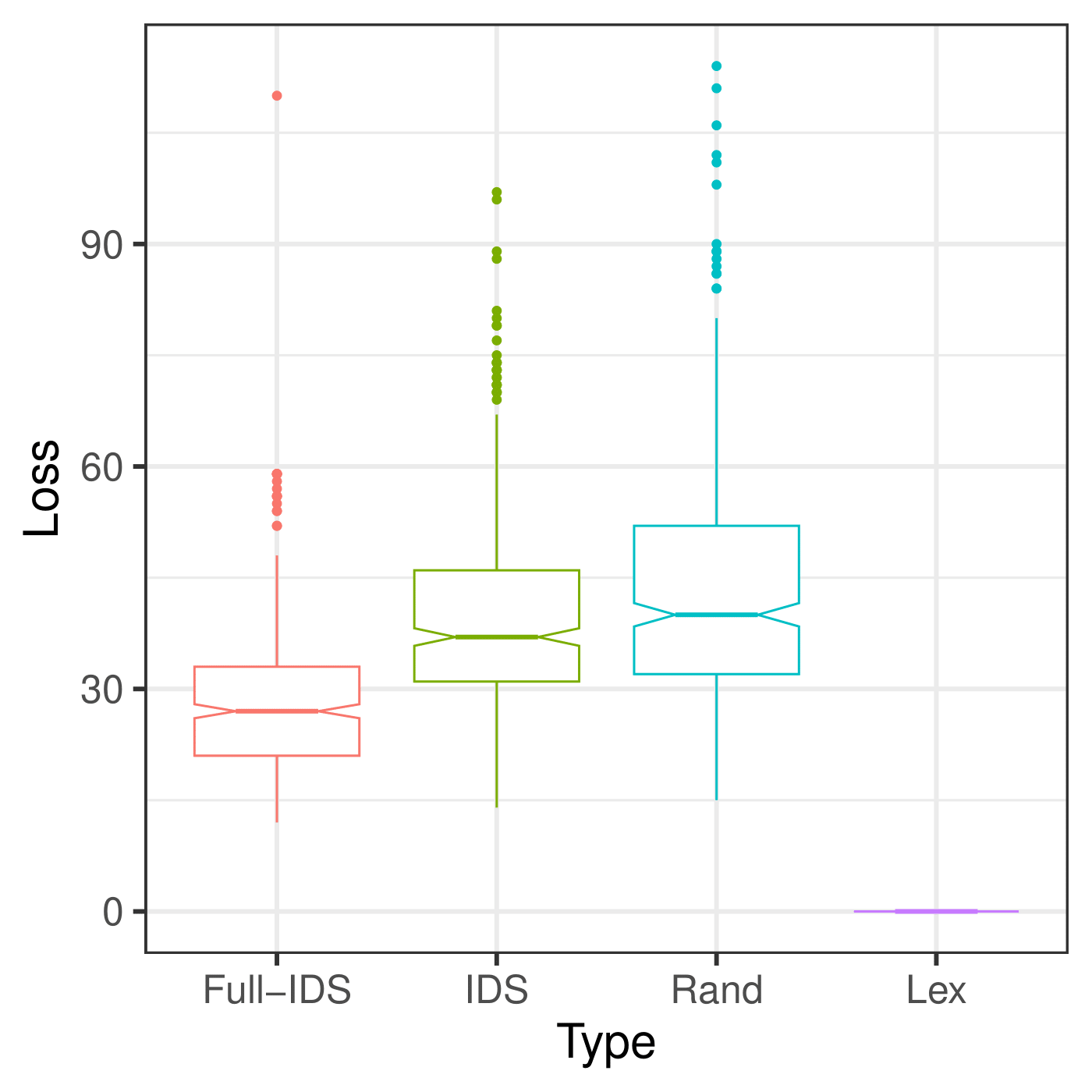}
         \label{fig:co005}
     \end{subfigure}
     \hfill
     \begin{subfigure}[b]{0.3\textwidth}
         \centering
         \caption{Count Odds, $r=0.1$}
         \includegraphics[width=\textwidth]{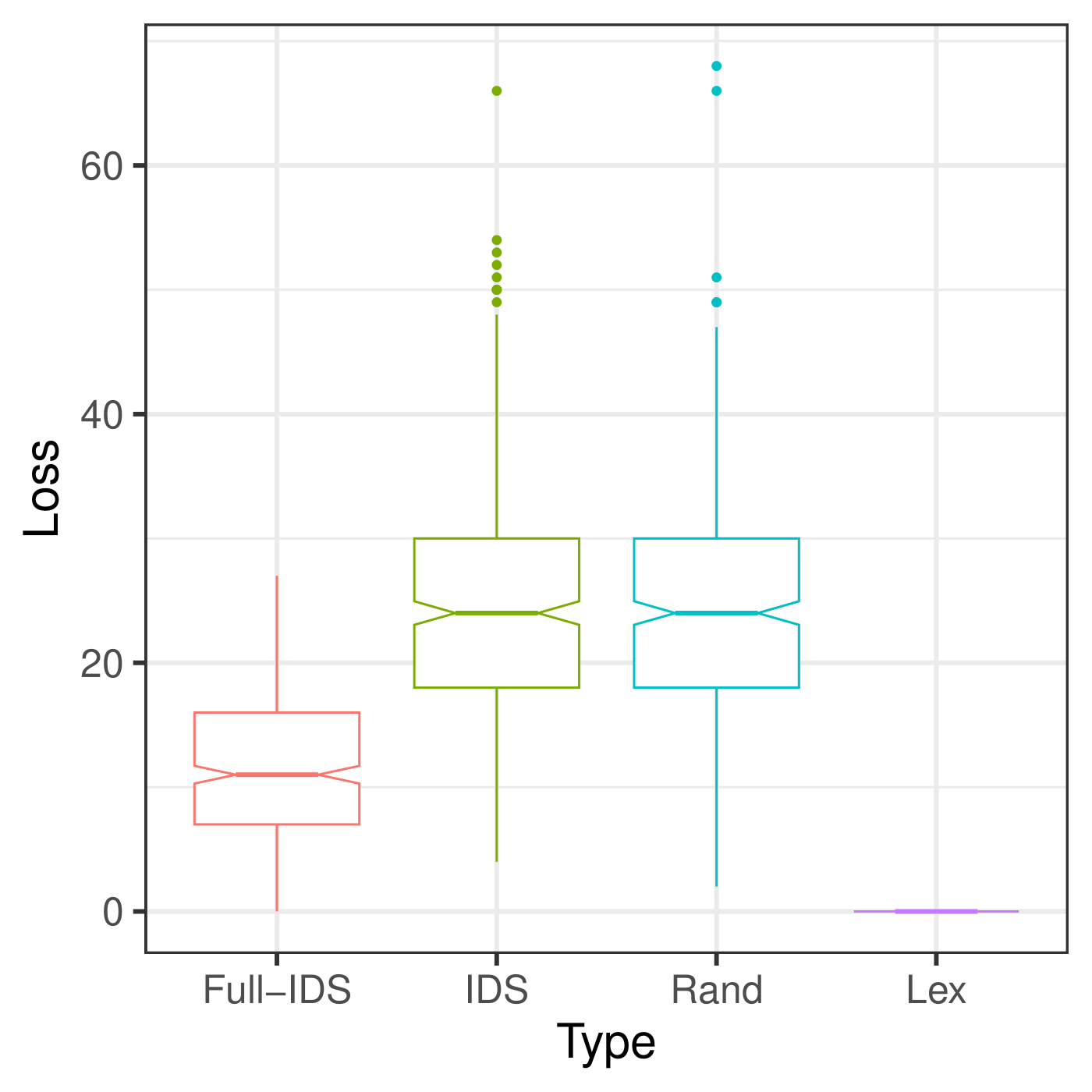}
         \label{fig:co01}
     \end{subfigure}
     \hfill
     \begin{subfigure}[b]{0.3\textwidth}
         \centering
         \caption{Count Odds, $r=0.2$}
         \includegraphics[width=\textwidth]{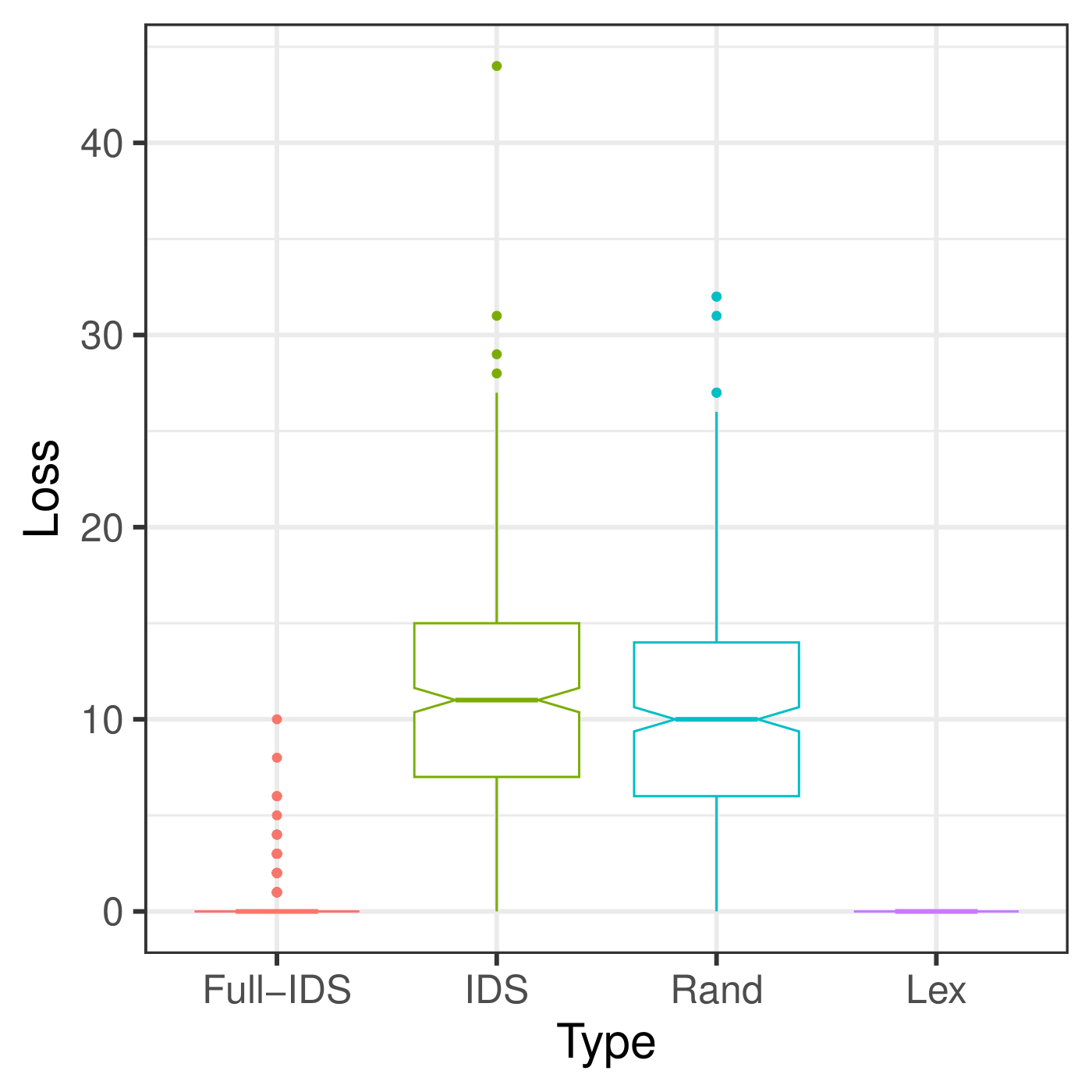}
         \label{fig:co02}
     \end{subfigure}
     \end{subfigure}
     \hfill
     \begin{subfigure}[b]{0.9\textwidth}
         \begin{subfigure}[b]{0.3\textwidth}
         \centering
         \caption{Fizz Buzz, $r=0.05$}
         \includegraphics[width=\textwidth]{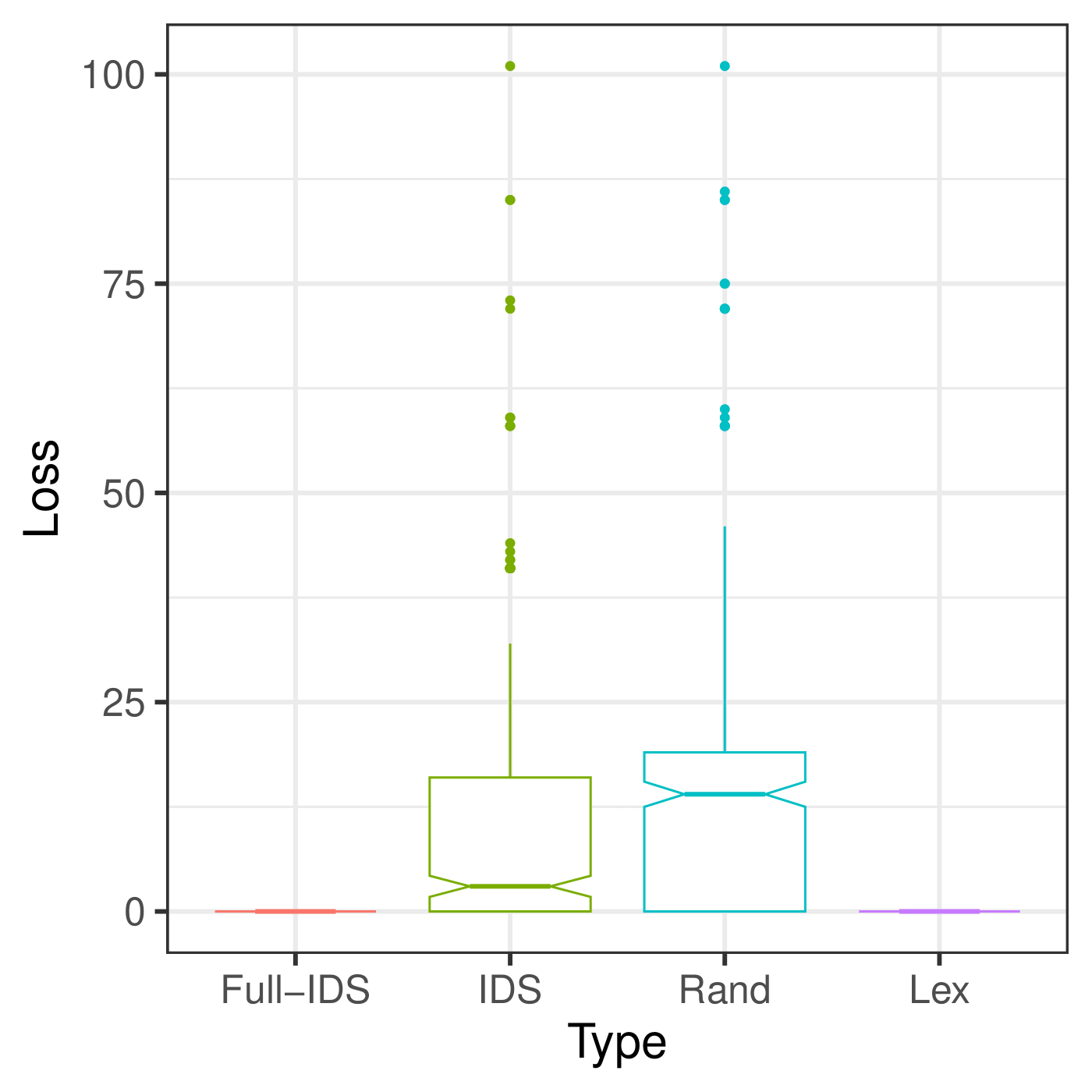}
         \label{fig:fb005}
     \end{subfigure}
     \hfill
     \begin{subfigure}[b]{0.3\textwidth}
         \centering
         \caption{Fizz Buzz, $r=0.1$}
         \includegraphics[width=\textwidth]{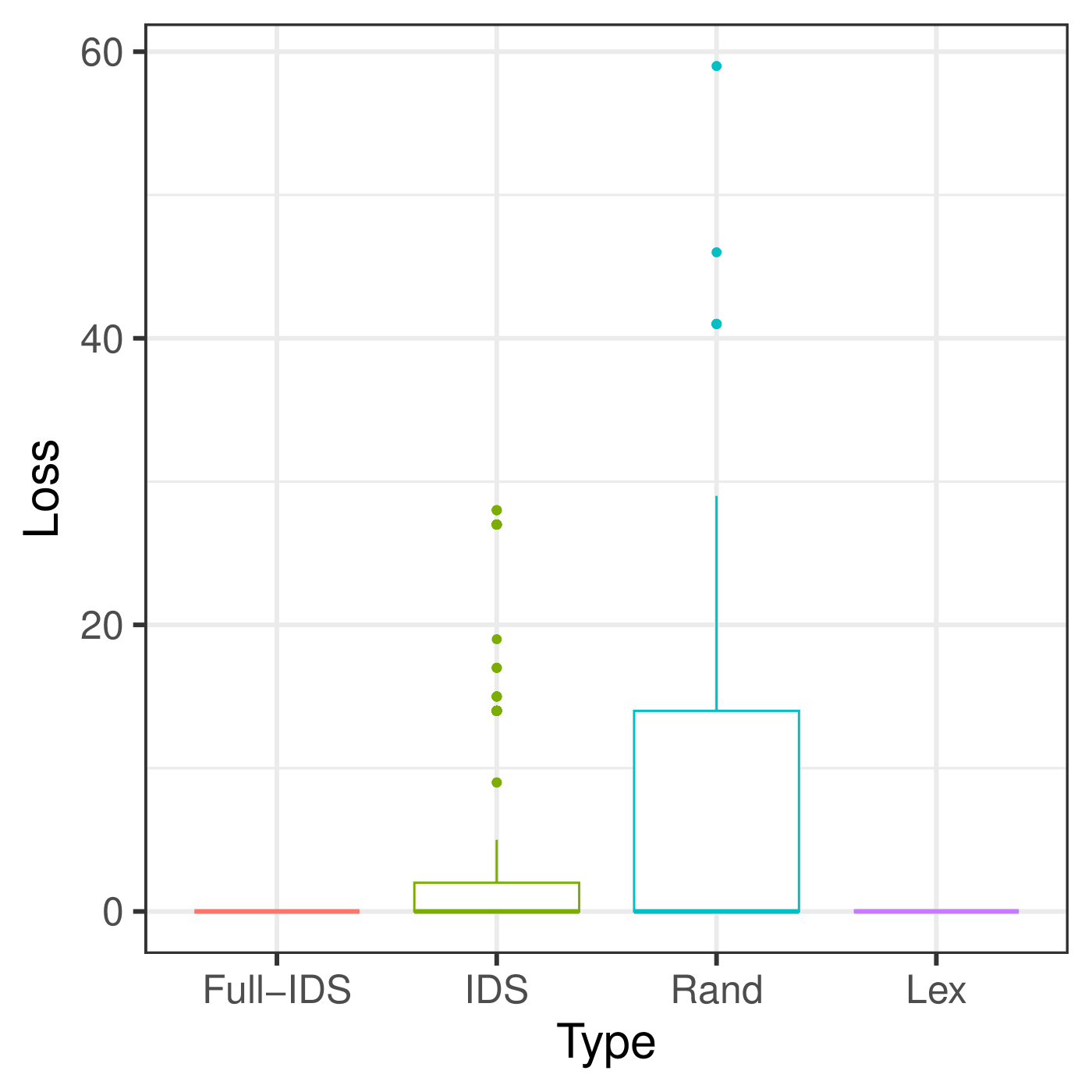} 
         \label{fig:fb01}
     \end{subfigure}
     \hfill
     \begin{subfigure}[b]{0.3\textwidth}
         \centering
         \caption{Fizz Buzz, $r=0.2$}
         \includegraphics[width=\textwidth]{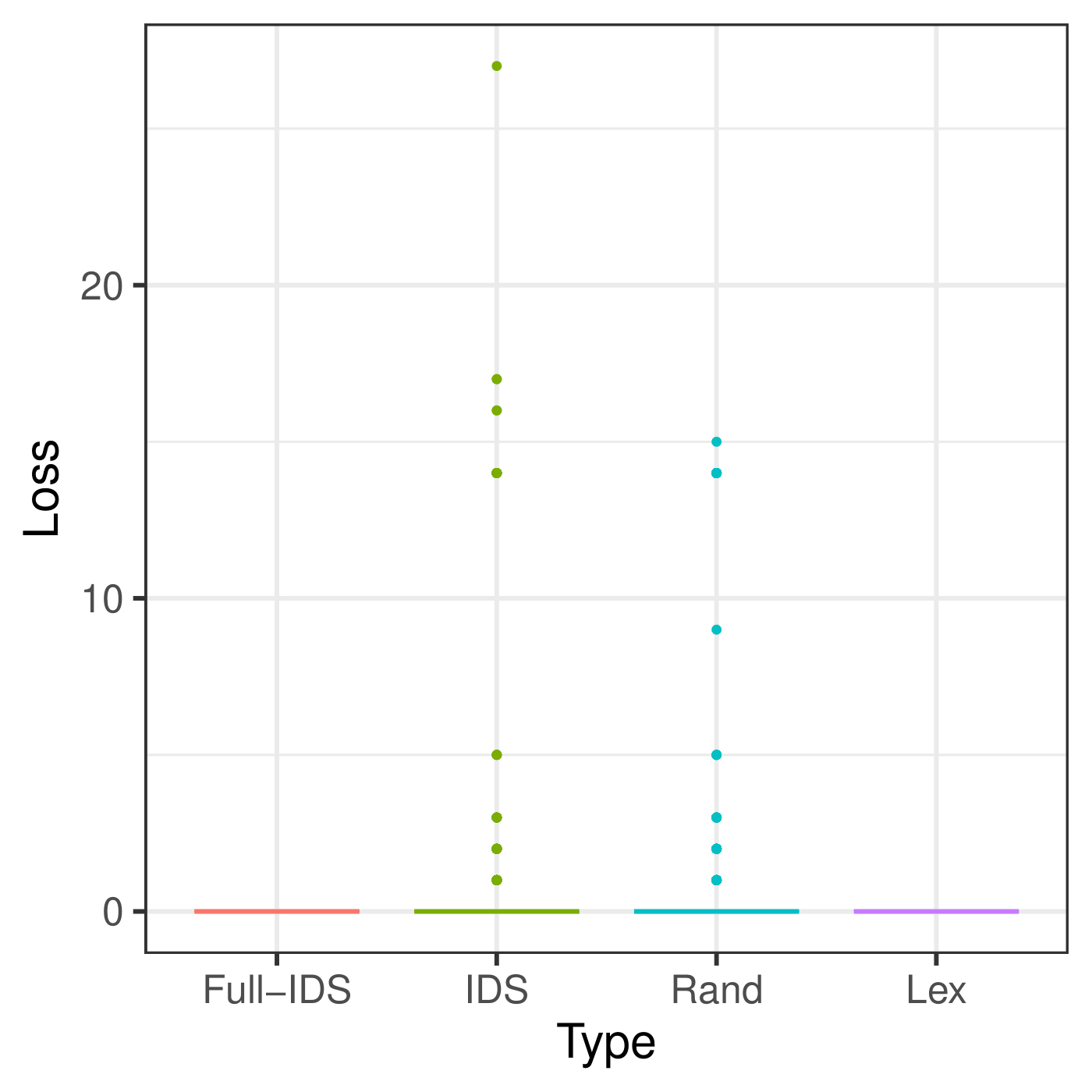}
         \label{fig:fb02}
     \end{subfigure}
     \end{subfigure}
        
        \label{fig:co}
\end{figure*}

%% file: figs/synthetic-pop-fig.tex
\begin{figure}[H]
    \centering
    \includegraphics[width=0.34\textwidth]{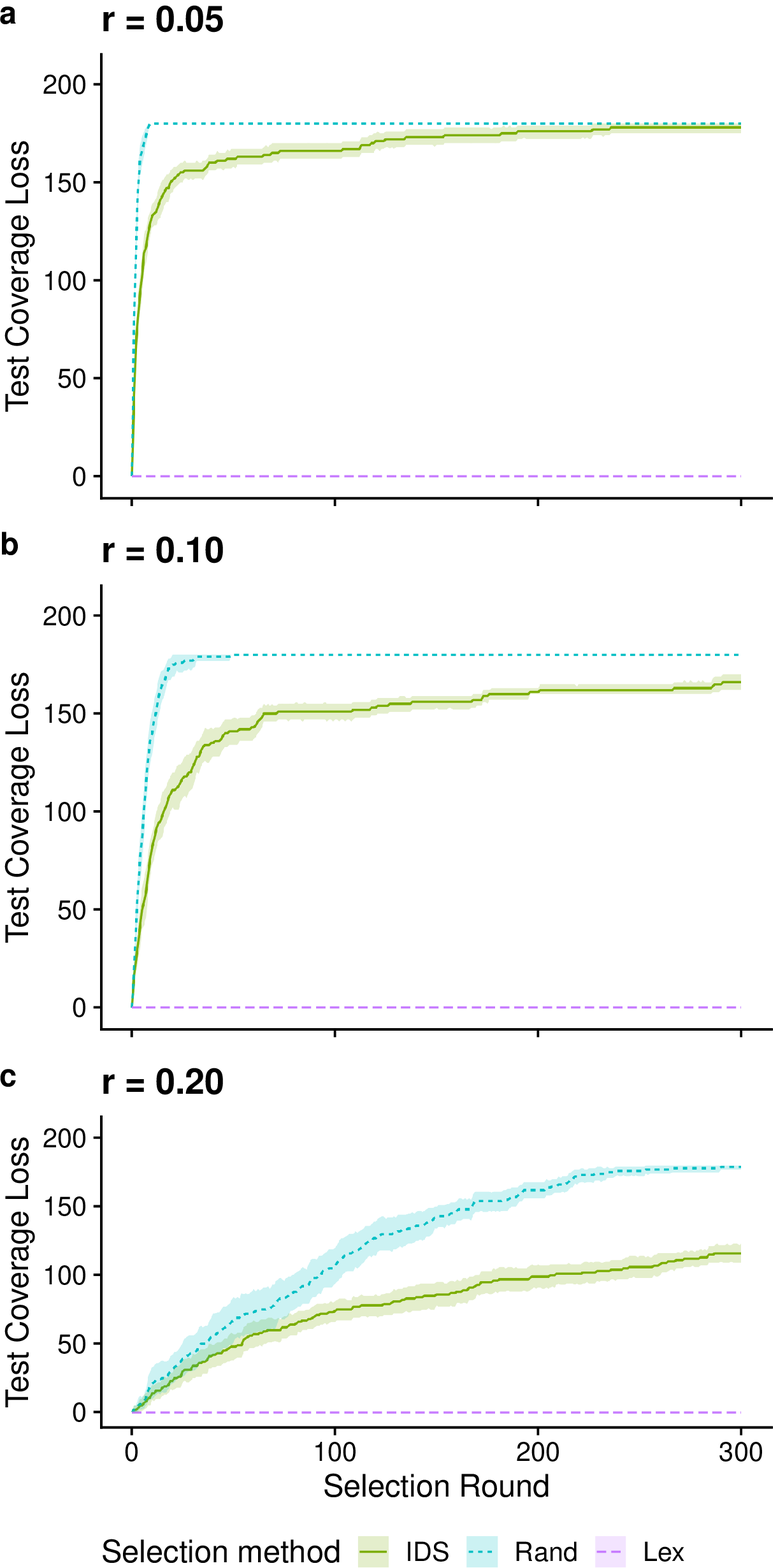}
    \caption{
    Synthetic population test coverage loss over 300 rounds of selection for down-sampling rates (a) 0.05, (b) 0.10, and (c) 0.20.
    Each line gives the mean value across 20 replicates, and the shading around each line indicates a 95\% confidence interval.
    }
    \label{fig:synthetic-pops}
\end{figure}